\documentclass[
twocolumn,
]{ceurart}

\usepackage{listings}
\usepackage{algorithm}
\usepackage{algpseudocode}
\begin{document}

\copyrightyear{2022}
\copyrightclause{Copyright for this paper by its authors.
  Use permitted under Creative Commons License Attribution 4.0
  International (CC BY 4.0).}

\conference{LLM4AI'23: Workshop on Foundations and Applications in Large-scale AI Models
-Pre-training, Fine-tuning, and Prompt-based Learning, co-located with the 29TH ACM SIGKDD CONFERENCE ON KNOWLEDGE DISCOVERY AND DATA MINING (KDD), August 6-10, 2023, Long Beach, CA, USA}

\title{AutoHint: Automatic Prompt Optimization with Hint Generation}

\author[1]{Hong Sun}[%
email=hosu@microsoft.com,
]

\author[1]{Xue Li}[%
email=xeli@microsoft.com,
]

\author[1]{Yinchuan Xu}[%
]

\author[1]{Youkow Homma}[%
]

\author[1]{Qi Cao}[%
]

\author[1]{Min Wu}[%
]

\author[1]{Jian Jiao}[%
]

\author[1]{Denis Charles}[%
]

\address[1]{Microsoft, 1045 La Avenida, Mountain View, CA, 94043, USA}




\begin{abstract}
  This paper presents \textbf{AutoHint}, a novel framework for automatic prompt engineering and optimization for Large Language Models (LLM). While LLMs have demonstrated remarkable ability in achieving high-quality annotation in various tasks, the key to applying this ability to specific tasks lies in developing high-quality prompts. Thus we propose a framework to inherit the merits of both zero-shot learning and few-shot learning by incorporating enriched instructions derived from input-output demonstrations to optimize original prompt. We refer to the enrichment as the \textit{Hint} and propose a framework to automatically generate the hint from labeled data. More concretely, starting from an initial prompt, our method first instructs a LLM to deduce new hints for selected samples from incorrect predictions, and then summarizes from per-sample hints and adds the results back to the initial prompt to form a new, enriched instruction. The proposed method is evaluated on the BIG-Bench Instruction Induction dataset for both zero-shot and few-shot prompts, where experiments demonstrate that our method is able to significantly boost accuracy for multiple tasks.
\end{abstract}


\maketitle

\section{Introduction}
\label{intro}

Large Language Models (LLM) have shown remarkable ability to achieve comparable or even surpass human annotation quality in various tasks~\cite{gilardi2023chatgpt}, where high-quality prompts are the key to activate such abilities. This has entailed explorations from many directions with a large body of studies on prompt engineering, including methods built upon human instinct or domain knowledge~\cite{wu2022ai}, data-driven approaches~\cite{zhou2022large}, or prompt optimization via in-context learning~\cite{zhang2023tempera,pryzant2023automatic}, etc.

\begin{figure*}[!t]
\begin{center}
   \includegraphics[width = 0.98\linewidth]{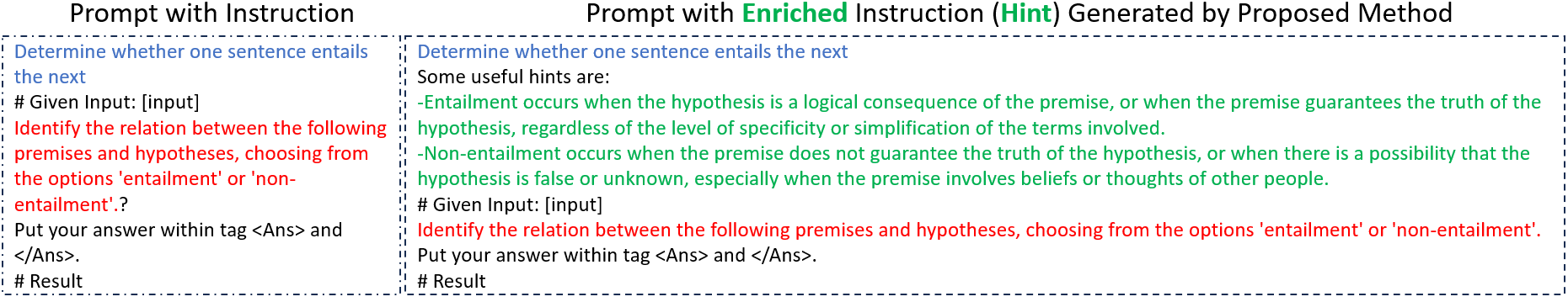}
\end{center}
   \caption{An example of the enriched instruction learned by our proposed method, taken from Epistemic Reasoning task in BBII dataset, with task description highlighted in blue, instruction in red and enriched hints in green. Left: a standard prompt with very general instruction. Right: the enriched prompt after adding the learned hints.
   }
   \label{fig.example}
\end{figure*}

LLMs could be prompted by zero-shot or few-shot learning. In the former, the LLM will be provided general instructions for the task at hand, 
as shown in the example in the left of Figure~\ref{fig.example}. By contrast, under a few-shot setting, LLM is provided with a number of input-output pairs as demonstrations followed by an unseen input, and then is instructed to generate a corresponding output. 
Both of these two approaches come with their own advantages and disadvantages: the former exhibits better generalization but may lack the necessary clarity and specificity, since the instructions can be vague or provide only general descriptions which is not easy for LLM to interpret, whereas the latter one offers more detailed information by providing demonstrations but is sensitive to sample selection or even sample ordering~\cite{lu-acl2022, xu2023reprompting, pryzant2023automatic}. 

The above analysis motivates us to seek an alternative approach by combining merits from both sides, via inducting enriched instructions from input-output demonstrations and then employing them to refine the original instruction. Throughout this paper, we will refer to these enriched instructions as \textbf{Hints}. 
Figure~\ref{fig.example} (right) shows an example of the hints generated for the \textit{Epistemic Reasoning} task in BIG-Bench Instruction Induction (BBII)~\cite{zhou2022large} dataset. The original description for this task is to \textit{Determine whether one sentence entails the next}, as highlighted in blue. In contrast, the hints generated by our method (highlighted in green) showcase a more elaborate instruction, providing a breakdown of explanations for both entailment and non-entailment cases. As a result, it provides more clear information for LLM to interpret the task and proceed with expected actions.

Having the enriched hints is good, but having automatically generated hints is better.
That is why we propose a framework called \textbf{AutoHint} which aims to generate hints \textbf{automatically}. Our framework samples from input-output pairs and leverages LLM for dueucing hints accordingly. The sampling is based on wrongly answered samples, as it is fair to assume that the original prompt has already conveyed sufficient information for the correctly answered ones. Following that, we select a small subset and prompt the LLM to summarize the hints that align best with that subset without including case-specific information. These hints are incorporated into the original prompt to generate a refined version. As our method is orthogonal to zero-shot and few-shot settings, we evaluate it under both settings on BBII dataset, where we observe remarkable accuracy improvement.

Our main contributions are three folds:
\begin{itemize}
    \item We propose \textbf{AutoHint}, an automatic prompt optimization framework which inherits merits from both zero-shot and few-shot learning. It adopts a residual-sampling-summarize paradigm to enhance robustness to noisy samples. 
    \item We report extensive evaluation results on BBII dataset to demonstrate the effectiveness of our method.
    \item Our framework explores fully leveraging GPT-4's capability for prompt optimization to reduce manual effort.
\end{itemize}


\section{Related Work}
\label{related}


Since the explosive arrival of LLMs~\cite{ouyang2022training}, many works have studied how to boost their capabilities to solve emergent, complex tasks, among which in-context learning and Chain-of-Thought\cite{wei2022chain} have proved to be effective two of the most effective methods. 
Along this line of research, different directions have been explored~\cite{cobbe2021training, wu2022ai, kojima2022large, wang2022self, zelikmanstar, zhou2022least, lu2022dynamic}. 
Nevertheless, in most tasks the most effective prompts are still human-crafted, which greatly limits the application of LLM prompting.

To fill this gap, there has been a growing interest in automatic prompt engineering, with most prior attempts requiring access to the internal variables of LLMs on different levels, either through the differentiable tuning of soft prompts \cite{qin2021learning, lester2021power} or training auxiliary components or models for prompt optimization\cite{deng2022rlprompt, hao2022optimizing, wang2022self, zhang2023tempera, shin2020autoprompt, long2023large}. 
However, with the growing trend of limited accessibility to LLMs, such methods are becoming less feasible for general practitioners. 


For that reason, recent works are witnessing a noticeable shift towards approaches that solely rely on feedback from LLMs \cite{zhou2022large, honovich2022instruction, pryzant2023automatic, xu2023reprompting}. More specifically, \cite{honovich2022instruction} instructs LLMs to deduce the task based on input-output demonstrations, and this idea is further extended into a generation-scoring-selection workflow in \cite{zhou2022large}. 
In contrast to existing works, our method enriches the general instructions obtained from \cite{zhou2022large, honovich2022instruction}, via exploiting complementary hints with enhanced clarity and specificity from input-output demonstrations.
Therefore, our proposed method is orthogonal to existing works and can be combined with them to obtain potentially better results.






\section{Proposed Method}
\label{method}

\begin{figure*}[!h]
\begin{center}
   \includegraphics[width = 0.7\linewidth]{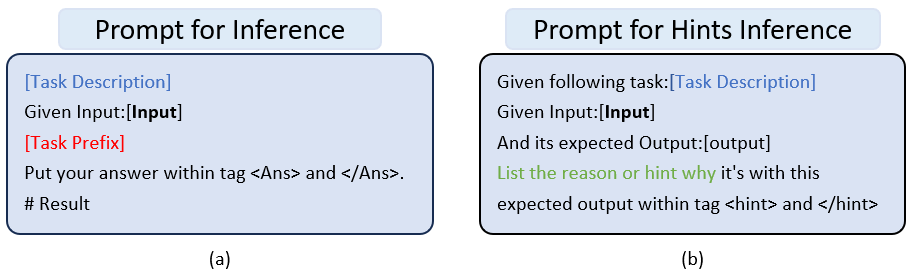}
\end{center}
   \caption{(a): Prompt template for inferencing output as in Section~\ref{method.inference}. (b): Prompt template for hints generation as in Section~\ref{method.hintsgen}.
   }
   \label{fig.p0}
\end{figure*}


\subsection{Overview}
\label{method.overview}

Given a task with a training dataset containing $N$ $\mathrm{i.i.d.}$ samples as input-output demonstrations: $\mathcal{D}_{train}=\{(x_1, y_1), \cdots, (x_N, y_N)\}$, our method first inferences the predictions with an initial prompt $p_{t}$, and then uses another prompt $p_h$ to deduce a reason or hint $r_i$ for each sample having incorrect predictions. We denote the wrongly predicted samples as $\mathcal{D}_{residual}=\{(x_1, y_1, r_1), \cdots, (x_n, y_n, r_n)\}$, and then sample $M$ samples from $\mathcal{D}_{residual}$ to form a subset $\mathcal{S}=\{(x_1, y_1, r_1),  \cdots, (x_M, y_M, r_M)\}$. After that, we leverage a prompt $p_s$ to summarize all the individual $r_i$ in $\mathcal{S}$ into a final hint $r'$. Finally, a new prompt is generated by merging learned $r'$ into $p_{t}$ to form prompt 
 $p_{t+1}$. The overall procedure is summarized in Algorithm \ref{algo.overall}.

\begin{algorithm*}
    \caption{AutoHint for Prompt Optimization}\label{algo.overall}
    
    \begin{algorithmic}[1]
        \Require Training set $\mathcal{D}_{train}=(x_i, y_i)$ with $i \in [1,N]$, validation set $\mathcal{D}_{val}$, test set $\mathcal{D}_{test}$, $p_h$, $p_s$, an initial prompt $p$.
        \State Initialize: $p_0=p$, $\mathcal{D}_{residual}=\emptyset$
        \For{$t=0, \cdots, T$}
            \State 
            $\hat{y}_i= p_{LLM}(x_i, p_{t}) $ for $i \in [1,N]$
           \State  $\mathcal{D}_{residual} =\{ (x_i, y_i, \hat{y}_i)\}\ for\ \hat{y}_i\ \neq y_i  $
           \State  $r_i \Leftarrow p_{LLM}(r_i | (x_i, y_i), p_{h})$ for $(x_i, y_i) \in \mathcal{D}_{residual}$ Section~\ref{method.hintsgen}
           \State $S \Leftarrow$ sampling($\mathcal{D}_{residual}$) Section~\ref{method.clustering}
             \State $r' \Leftarrow p_{LLM}( (x_i, y_i, r_i), p_{s})$ Section~\ref{method.summarization}
             \State Generate $p_{t+1}$ based on $p_t$ and $r'$
        \EndFor  
    \end{algorithmic}
\end{algorithm*}

    


\subsection{Inference and Get Residual Data}
\label{method.inference}
Figure~\ref{fig.p0}(a) shows the template used to inference prediction $\hat{y}_i$ for each sampled input data from $\mathcal{D}_{train}$:
\begin{eqnarray}\label{eqn.inference}
p_{LLM}(\hat{y}_i | (x_i), p_{t})
\end{eqnarray}
where $p_{t}$ is the initial prompt constructed by concatenating a very general task description with the sampled input.


Upon obtaining the inference results, we will only retain incorrectly predicted samples i.e., $\mathcal{D}_{residual}$, 
 for further processing. The reasons for this are twofold. First, proceeding with all training data would result in significant costs. Second, we believe that the current prompt already provides sufficient information for the LLM to make accurate decisions for correctly predicted data. 
Therefore, we only send the error cases to subsequent steps, aiming to generate new information that can assist in rectifying the current decision.


\subsection{Hints Generation}
\label{method.hintsgen}
After having the residual dataset, another prompt is used to generate $r_i$ for each input-output pair:
\begin{eqnarray}\label{eqn.inference}
r_i \Leftarrow p_{LLM}(r_i | (x_i, y_i), p_{h})
\end{eqnarray}
where $p_{h}$ is a template prompt constructed by concatenating the general task description and the corresponding input-output pair, as shown in Figure \ref{fig.p0}(b). 


\subsection{Sampling Strategies}
\label{method.clustering}

We explore various sampling strategies to generate $S$, including
\begin{itemize}
    \item \textit{Random}: randomly draw $K$ samples from $\mathcal{D}_{residual}$.
    \item \textit{Random-balanced}: randomly draw $K$ samples from each category (applies to classification tasks only).
    \item \textit{Clustering}: first cluster samples based on $x_i, y_i, r_i$, and then randomly draw $K$ samples from each cluster. 
\end{itemize}

Our framework makes no assumptions on specific clustering methods for \textit{Clustering} strategy, and we employ K-Means in our implementation to obtain the cluster index $c_i$ for sample $i$ based on its feature vector $v_i$
for $\forall i \in S$, due to its simplicity:
\begin{eqnarray}\label{eqn.kmeans_vec}
v_i = \lambda_{x} f(x_i) + \lambda_{y} f(y_i) + \lambda_{r} f(r_i),
\end{eqnarray}
where $f(\cdot)$ denotes the encoder that transforms the textual input $x_i$, $y_i$ and $r_i$ into dense vectors, with $\lambda_{x}$, $\lambda_{y}$ and $\lambda_{r}$ denoting the combination weights. We use the BERT-base model as our encoder $f(\cdot)$, but other encoders could be easily integrated into our framework.


As will be discussed in Section~\ref{exp}, we use a validation set to examine different sampling strategies and select the best performing one, and then report its evaluation results on the holdout test set.





\subsection{Hints Summarization}
\label{method.summarization}

\begin{figure}[!h]
\begin{center}
   \includegraphics[width = 0.8\linewidth]{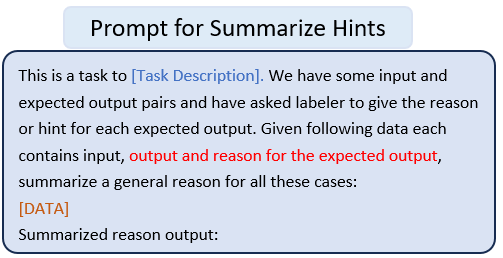}
\end{center}
   \caption{Prompt template for summarizing hints.  
   }
   \label{fig.ps}
\end{figure}

Given the hints generated from Section~\ref{method.hintsgen}, we leverage another prompt $p_{s}$ shown in Figure \ref{fig.ps} to generate a summarized hint to add to the initial prompt $p_{t}$, resulting in an enriched prompt $p_{t+1}$. An example of the enriched prompt with the summarized hints is shown in Figure~\ref{fig.example} (right). 

The idea of summarizing individual hints here is conceptually analogous to the calculation involved in deriving a mini-batch stochastic gradient descent (SGD), where individual gradients within the mini-batch are aggregated into a single gradient. However, compared to its numerical counterpart, summarizing hints is much more difficult, since the former operates in a continuous space with well-established numerical operations, which is clearly not the case for hints summarization. In this paper, we propose to transform this non-trivial task into a tractable problem by leveraging the ability of LLMs to solve emergent tasks. To our knowledge, this is the first attempt to leverage LLMs to summarize hints or prompts in automatic prompt generation to make the result more stable and comprehensive.


    






\section{Experiments}
\label{exp}



\subsection{Experimental Setup}
\label{exp.setup}

\textbf{Dataset} We examine our method on 6 tasks from the BIG-Bench Instruction Induction (BBII)\footnote{\url{https://github.com/keirp/automatic_prompt_engineer/tree/main/data/bigbench-ii}} dataset introduced in \cite{zhou2022large}, including \textit{Epistemic Reasoning}, \textit{Logical Fallacy Detection}, \textit{Implicatures}, \textit{Hyperbaton}, \textit{Causal Judgment} and \textit{Winowhy}. 
We specifically select these tasks based on their relatively large data volume and easy-to-parse answer format, which allows us to obtain higher accuracy.
For each task, we randomly split the dataset into training (60\%), validation (20\%) and test sets (20\%), where the training set is used for inducing the hints as mentioned in Section~\ref{method}, while the validation set is used for hyper-parameter tuning and prompt selection. After choosing the best enriched prompt based on the validation set performance, we report the final evaluations on the test set.

\textbf{Baselines and Metrics}
Since our proposed framework aims to exploit useful information on top of given prompts, we employ the initial prompt $p_{t}$ as our baseline, which is the prompt retrieved from BBII dataset in our implementation. Given all the tasks are binary-classification tasks, we report Overall Accuracy (Acc.) and Balanced Accuracy (Bal Acc.) as our main evaluation metrics. Balanced Accuracy is averaged per-category accuracy.

\textbf{Implementation details}
We use the Azure OpenAI API service (GPT-4) to evaluate both the baseline and our method. For the inference and summarization prompts, we set temperature as 0 and topP as 1. For hint generation, we set temperature as 0.1 and topP as 0.95 to allow some exploration. We conduct hyper-parameter tuning to examine different sampling strategies. For each task, we test 20 sets of hyper-parameters on the validation set and select the best one to conduct evaluation on the test set. 
To save costs, we randomly sample a subset from the validation set for comparison if it is too large, then keep the top three settings to run a full validation set evaluation and select the final prompt for the test set evaluation. 

\begin{table*}
  \centering
  \caption{Zero-Shot Prompt Evaluation Results}
  \label{tab:eval}
  \begin{tabular}{p{0.3\linewidth} p{0.1\linewidth} p{0.1\linewidth} p{0.1\linewidth} p{0.1\linewidth} p{0.1\linewidth}}
    \toprule
    Task & Testset  & \multicolumn{2}{c}{Baseline} & \multicolumn{2}{c}{Treatment}\\
   & Count &  Acc. & Bal Acc. & Acc. & Bal Acc. \\
    \midrule
   Epistemic Reasoning & 499 & 82.5 & 82.9 & \textbf{90.5} & \textbf{90.15} \\
    Logical Fallacy Detection & 700 & 77.46 & 76.18 & \textbf{84.03} & \textbf{83.66} \\
     Implicatures & 100 & 88.89 & 88.66 & \textbf{91.93} & \textbf{91.92} \\
      Hyperbaton & 11k & 66.81 & 66.81 & \textbf{82.41} & \textbf{82.41} \\
      Causal Judgment & 40 & 60.53 & 60.53 & \textbf{65.79} & \textbf{65.79} \\
       Winowhy &570 & 74.17 & 73.24 & 73.12 & 71.78 \\
  \bottomrule
\end{tabular}
\end{table*}
\subsection{Experiment Results and Analysis}
\label{exp.overall}
Our main results are summarized in Table~\ref{tab:eval}, where we observe significant improvement on both accuracy and balanced accuracy on 5 out of the 6 tasks after adding the hints mined from the proposed method, demonstrating the effectiveness of our proposed framework under zero-shot settings.

We also look into the generated hints and observe that for most tasks, our method is able to mine useful definitions and explanations with more detailed information after the very first iteration. For the example in Figure~\ref{fig.example}, practical instructions are added for both the entailment and non-entailment segments, assisting annotators including LLMs to better understand the task at hand. 

Regarding sampling strategies, \textit{Clustering} achieves the best result in 3 
out of 6 tasks, \textit{Random-balanced} leads to the best result on the remaining tasks, while plain \textit{Random} sampling shows the worst results. 
This highlights the importance of adopting proper sampling strategies for generating a good summary. 
Meanwhile, the remarkable performance of \textit{Random-balanced} sampling also echos our finding that our method is able to induce explicit per-category explanations as shown in Figure~\ref{fig.example}. In addition, the best results are achieved when no more than 3 samples are used per label.  This may be because more samples confuse the model when generating the summary. 



\begin{table*}
  \centering
  \caption{Few-Shot Prompt Evaluation Results}
  \label{tab:few}
  \begin{tabular}{p{0.4\linewidth}  p{0.1\linewidth} p{0.1\linewidth} p{0.1\linewidth} p{0.1\linewidth}}
    \toprule
    Task & \multicolumn{2}{c}{Baseline} & \multicolumn{2}{c}{Treatment}\\
   &   Acc. & Bal Acc. & Acc. & Bal Acc. \\
    \midrule
   Epistemic Reasoning &  76.5 & 80.87 & \textbf{83.75} & \textbf{85.96} \\
    Logical Fallacy Detection &  86.76  & 86.64 & 84.64 & 84.28 \\
     Implicatures & 88.89 & 88.73 & \textbf{92.93} & \textbf{92.89} \\
      Hyperbaton &  61.88 & 61.88 & \textbf{80.41} & \textbf{80.41} \\
      Causal Judgment & 55.26 &  55.26 & \textbf{73.68} & \textbf{73.68} \\
       Winowhy & 71.55 & 70.43 & \textbf{74.69} & \textbf{74.04} \\
  \bottomrule
\end{tabular}
\end{table*}

\begin{table*}
  \centering
  \caption{Accuracy comparison after second iteration (zero-shot)}
  \label{tab:second_it}
  \begin{tabular}{p{0.4\linewidth}  p{0.1\linewidth} p{0.1\linewidth} p{0.1\linewidth} p{0.1\linewidth}}
    \toprule
    Task & \multicolumn{2}{c}{First iteration} & \multicolumn{2}{c}{Second iteration}\\
   &   Acc. & Bal Acc. & Acc. & Bal Acc. \\
    \midrule
   Epistemic Reasoning       & \textbf{90.5}  & \textbf{90.15} & 88.2           & 88.2 \\
   Logical Fallacy Detection & 84.03          & 83.66          & \textbf{87.68} & \textbf{87} \\
   Implicatures              & 91.93          & 91.92          & \textbf{92.92} & \textbf{91.91} \\
   Hyperbaton                & 82.41          & 82.41          & \textbf{90.56} & \textbf{90.56} \\
   Causal Judgment           & 65.79          & 65.79          & \textbf{71.05} & \textbf{71.05} \\
   Winowhy                   & 73.12          & 71.78          & 73.1           & 71.8 \\
  \bottomrule
\end{tabular}
\end{table*}


Meanwhile, as our method is orthogonal to few-short learning, we also evaluate AutoHint in this setting, and the results are summarized in Table~\ref{tab:few}, where our method boosts accuracy significantly in 5 out of 6 tasks. After adding demonstrations randomly sampled from the training set, we observe a decrease in accuracy on 4 out of 6 tasks, along with notable performance fluctuation across different sampling seeds. This observation aligns with previous studies~\cite{lu-acl2022,xu2023reprompting} 
highlighting the importance of carefully selecting effective demonstrations to mitigate the sensitivity towards samples.


\subsection{More Iterations}
\label{exp.iterations}

In addition, we also conduct experiments by running additional iterations to generate more informative prompts. New hints are appended to existing ones by a prefix \emph{And additional hint might be useful is}. Overall, we observe a significant boost on the average accuracy over prompt candidates on the validation set, this indicates our improvement on the original prompt is solid. However, after selecting the best prompt, only 3 tasks (\textit{Causal Judgement}, \textit{Logical Fallacy Detection} and \textit{Hyperbaton}) show further improvement while the rest show accuracy decay. When manually checking the new hints, we find that they indeed present complementary or more specific information to existing ones. For instance, new hints for the \textit{Epistemic Reasoning} task focus on if the premise and hypothesis are identical or include additional details. 
While it is beneficial to infuse new, correct information, it places higher demands on the ability of LLM to comprehend these diverse hints. Our current approach to incorporating hints from later iterations is simplistic, and as a result, it hinders the language model from effectively assimilating the more informative instructions. As a result, it requires a more effective strategy for merging information from different iterations. Therefore, we plan to address this as part of our future work.

\subsection{Cost Analysis}
\label{exp.cost}
There are 4 calls to LLMs in each iteration: 1) inference, 2) hints generation, 3) summarization and 4) prompt selection. 
Among these, the first call is not unique to our method therefore it does not incur additional cost.
Given the cost for the third call is negligible, we only need to optimize costs for steps 2 and 4. For step 2, we can sample a subset first as the final data points needed to obtain the summary is small. For the last step, we can evaluate on a subset if the validation set is large, or similar to APE~\cite{zhou2022large}, iteratively prune the prompts.

\section{Conclusion}
\label{conclusion}

We propose a novel framework for automatic prompt optimization by combining merits from zero-shot and few-shot learning. We evaluate our method on the BBII dataset under both settings, and demonstrate the effectiveness of the proposed method. Our proposed method could also be combined with existing methods to achieve potentially better results. While we didn't deploy this method in our production system yet, we are able to leverage this method for reducing some manual analysis effort in prompt optimization. There are more discussions along this work like more-representative sample selection, merging information from different iterations to generate better prompt and cost saving, which we leave them for future work.


\bibliography{main}




\end{document}